%% file: main.tex
\documentclass[conference]{IEEEtran}
\IEEEoverridecommandlockouts

\usepackage{cite}
\usepackage{amsmath,amssymb,amsfonts,amsthm}

\usepackage{graphicx}
\usepackage{textcomp}
\usepackage{xcolor}
\usepackage{booktabs}
\usepackage{bm}
\usepackage{url}
\usepackage{multirow}
\usepackage{mathtools}

\usepackage[hidelinks,pdftex]{hyperref}

\usepackage[linesnumbered,ruled]{algorithm2e}
\SetKwInput{KwInput}{Input}
\SetKwInput{KwOutput}{Output}
\SetAlFnt{\small}
\SetAlCapFnt{\small}
\SetAlCapNameFnt{\small}

\makeatletter
\newcommand{\removelatexerror}{\let\@latex@error\@gobble}
\makeatother

\def\BibTeX{{\rm B\kern-.05em{\sc i\kern-.025em b}\kern-.08em
    T\kern-.1667em\lower.7ex\hbox{E}\kern-.125emX}}
\begin{document}

\title{One-Shot Domain Incremental Learning}

\author{
\IEEEauthorblockN{Yasushi Esaki$^\dag$ \hspace{3cm} Satoshi Koide$^\dag$ \hspace{3cm} Takuro Kutsuna$^\dag$}
\vspace{0.5cm}
\IEEEauthorblockA{\textit{$^\dag$Toyota Central R\&D Labs., Inc.} \\
Aichi, Japan \\
\texttt{\href{mailto:yasushi.esaki.sb@mosk.tytlabs.co.jp}{\nolinkurl{{yasushi.esaki.sb, koide, kutsuna}@mosk.tytlabs.co.jp}}}
}
}

\maketitle

\begin{abstract}
Domain incremental learning (DIL) has been discussed in previous studies on deep neural network models for classification. In DIL, we assume that samples on new domains are observed over time. The models must classify inputs on all domains. In practice, however, we may encounter a situation where we need to perform DIL under the constraint that the samples on the new domain are observed only infrequently. Therefore, in this study, we consider the extreme case where we have only one sample from the new domain, which we call one-shot DIL. We first empirically show that existing DIL methods do not work well in one-shot DIL. We have analyzed the reason for this failure through various investigations. According to our analysis, we clarify that the difficulty of one-shot DIL is caused by the statistics in the batch normalization layers. Therefore, we propose a technique regarding these statistics and demonstrate the effectiveness of our technique through experiments on open datasets. The code is available at \url{https://github.com/ToyotaCRDL/OneShotDIL}.
\end{abstract}


\section{Introduction}
\label{secintro}
In recent years, deep learning has been widely used in image recognition, speech recognition, and natural language processing~\cite{DeepLearning}.
There is a need to update a trained neural network model so that the model can classify samples correctly on a new, untrained input distribution (domain)~\cite{Continual}.
The additional training for the new domain is called domain incremental learning (DIL)~\cite{GenerativeReplay}.
In DIL, we need to correctly classify inputs on both the new and original domains.

\begin{figure}[t]
    \centerline{\includegraphics[height=4.3cm,width=7.5cm]{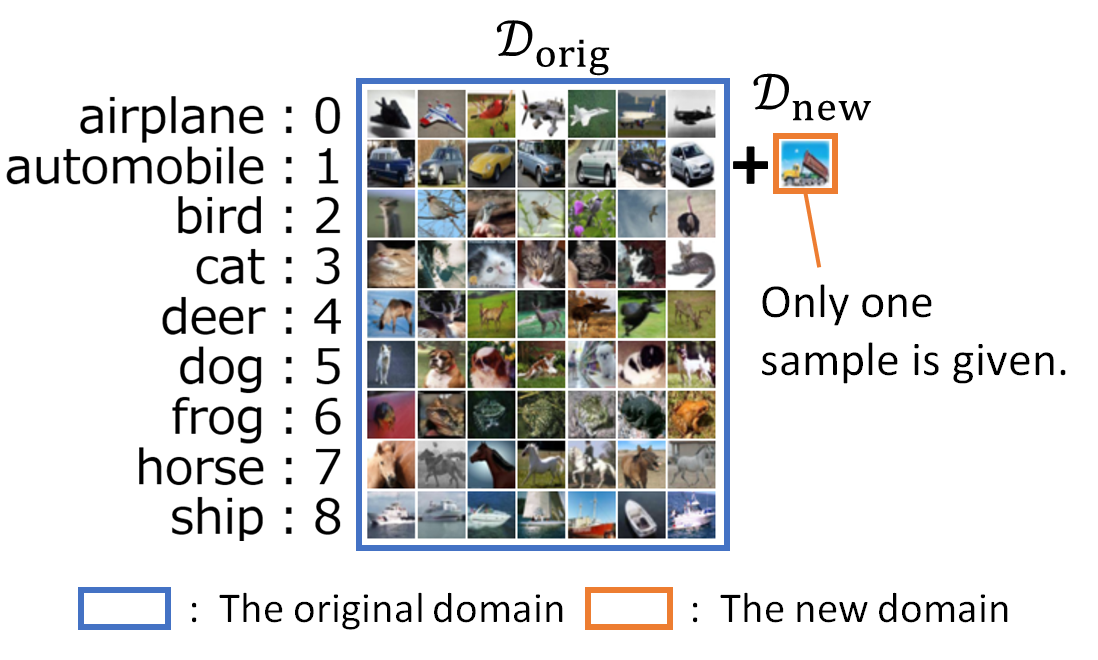}}
    \caption{Example of the original domain and the new domain in one-shot domain incremental learning (one-shot DIL) with CIFAR10~\cite{CIFAR}. In this example, trucks are added to the ``automobile'' class as the new domain. However, only one sample is added. The procedure to set datasets in one-shot DIL is described in Section~\ref{secdataset}.}
    \label{fig}
    \vspace{-0.3cm}
\end{figure}

Although previous studies on DIL assume many samples from the new domain, such an assumption sometimes does not hold in practical situations. 
For example, samples from rare domains are hardly obtained in the training dataset. In this case, the size of new domain samples available for DIL is also expected to be small, making the above assumption invalid.
For instance, consider the traffic sign recognition requirements of self-driving cars.
For classifying traffic signs into categories such as stop, one-way, and speed limit, based on images obtained from cameras, automatic classification is made possible by using neural network models trained on a dataset containing images of traffic signs~\cite{sign1,sign2}.
However, even if we train a large image dataset of traffic signs for this classification, we may encounter a new, unprecedented type of traffic sign that will be misclassified by the trained model in the test drives.
In the category of stop signs, many different types of stop signs are in use~\cite{sign3}. Each time we find a new type of stop sign, we must update the trained model.
In particular, even if there is only one stop sign in the world that looks like the one we found, we still need to update the model so long as the self-driving cars are using it.
In addition, the updated model is expected to successfully recognize the same stop sign in different situations.
As in this example, even when there is only one sample from the new domain, we need DIL techniques.
Therefore, in this study, we formulate a one-shot DIL and explore it.
Figure~\ref{fig} shows an example of the new domain and the original domain arranged using CIFAR10~\cite{CIFAR}, in one-shot DIL.
In this example, no truck images are included in the original nine-class data, and such images are not used in the first training.
Suppose that one truck image is given and added to the automobile class as a new domain after the first training.
We assume that the model obtained by the first training misclassifies the given truck image.
Therefore, we attempt to update this model with the given truck image as the second training, so that the model classifies truck images into the automobile class. 

Our research question is whether existing DIL methods also work in one-shot DIL, or not.
As we will see in the experiments, Elastic Weight Consolidation (EWC)~\cite{EWC} and Gradient Episodic Memory (GEM)~\cite{GEM}, which are typical DIL methods, do not work in one-shot DIL.
This result means that one-shot DIL is a more challenging problem than the general DIL setting since EWC and GEM achieve high accuracy in the general DIL setting. Therefore, a discussion specific to one-shot DIL is needed.

To address the difficulty in one-shot DIL, we carefully investigate the issues.
In this paper, we clarify the issues with one-shot DIL as being in the statistics of the batch normalization layers~\cite{BatchNorm}. 
In one-shot DIL, the moving averages of these statistics show unintended transitions not seen in general DIL, which negatively affects training and inference. Based on this fact, we propose a technique regarding these statistics to make the existing DIL methods work in one-shot DIL. We report experimental results indicating that our technique contributes to achieving high accuracy in one-shot DIL. Our technique does not interfere with other DIL methods and can be combined with various DIL methods. Therefore, this study plays the role of a baseline for one-shot DIL.

In summary, our contributions are as follows:
\begin{itemize}
    \item We examine one-shot DIL and show that existing DIL methods do not work in one-shot DIL due to the statistics in the batch normalization layers.
\item We propose a technique regarding the statistics in the batch normalization layers to make the existing DIL methods work in one-shot DIL and present a baseline for one-shot DIL.
\end{itemize}
\section{Related Work}
\subsection{Continual Learning}
DIL is a part of continual learning~\cite{Continual}.
There are three types of scenarios in continual learning: DIL, class incremental learning (CIL), and task incremental learning (TIL)~\cite{GenerativeReplay}.
In DIL, several datasets with different domains are observed over time.
All the datasets have the same number of classes, and a new domain is added to each class as a new dataset is observed. Therefore, the number of classes is constant.
Both the original and new domains must be classifiable by the training model.
In CIL, the number of classes increases with each new dataset added as new classes~\cite{CIL1,CIL2}.
Therefore, when a new dataset is observed, the number of output dimensions of the training model is increased.
The training model must classify all classes.
In TIL, the output spaces of all the datasets observed over time are disjoint~\cite{TIL1,TIL2}.
For example, consider the case where the original dataset deals with a 10-class classification problem, while the new dataset deals with a single output regression task~\cite{re-eval}.
Neural network models often have a multi-head output layer in TIL, wherein each head is responsible for a specific task.

The effect of batch normalization on continual learning has been studied outside of the one-shot case. Lange et al.~\cite{continualnorm} normalize not only the sample axis but also the channel axis to prevent the batch normalization layers from specializing in each domain. In this paper, we use a different approach to avoid the degradation of accuracy due to batch normalization.
\subsection{Few-Shot Continual Learning}
Recent studies on computer vision discussed few-shot CIL (FCIL)~\cite{FCIL2,FCIL3}.
FCIL is similar to CIL except that it involves a limited number of training samples.
In many cases, we assume that the first training dataset has a large sample size, while the subsequent datasets have smaller sizes~\cite{FCIL1}. A training model must discriminate between the new and original classes, even though the new classes have few samples.

Another study explored few-shot DIL~\cite{CFSL}.
The few-shot DIL called New Classes with Overwrite Settings in~\cite{CFSL} is similar to ours.
The difference lies in the number of classes in the new domain.
In the earlier work~\cite{CFSL}, the number of classes in the new domain is the same as that in the original domain.
In contrast, in the present work, the new domain contains only one class regardless of the number of classes in the original domain, and only one sample belongs to that one class.
The meta-learning methods used in~\cite{CFSL} assume that there are multiple classes and multiple samples in each domain.
The Self-Critique and Adapt model (SCA)~\cite{SCA}, which can achieve the best accuracy in~\cite{CFSL}, needs two datasets―a support set and a target set―for each domain.
The two datasets must be different because the target set is used for validation.
Hence, it is difficult to apply this method to our one-shot DIL.

\subsection{Neural Network Software Repair}
In our settings, a new sample from the new domain is assumed to be misclassified before training the new domain.
Therefore, we consider one-shot DIL as the process of repairing or debugging neural network models.
Some studies have discussed neural network software repair in software engineering~\cite{Apricot,Arachne}.
These studies aimed to reduce the number of misclassified samples by updating the weight parameters of a trained model.
Note that the predictions of correctly classified samples cannot be changed when the weight parameters are updated.
Some studies assumed 1000 samples as the number of samples to be repaired~\cite{DeepRepair,Few-Shot-Guided}. Other studies set the number of samples to be repaired depending on the accuracy of a pre-trained model~\cite{Apricot,SENSEI}.
Few studies discussed neural network software repair with limited training samples.
In contrast, the present study considers the situation of only one additional class with only one sample.
Therefore, our study is valuable not only for continual learning but also for neural network software repair.

\subsection{Transfer Learning}
Some studies on transfer learning discussed batch normalization.
Kanavati and Tsuneki~\cite{BN1} claimed that fine-tuning only the trainable weights and biases of the batch normalization layers yields performance similar to that achieved by fine-tuning all the weights of the neural network models.
Meanwhile, Yazdanpanah et al.~\cite{BN2} showed that shifting and scaling normalized features by the weights and biases of the batch normalization layers is detrimental in few-shot transfer learning.
They demonstrated that the removal of batch normalization weights and biases can have a positive impact on performance.
These studies focus on the weights and biases, while our study focuses on the statistics.
\subsection{Domain Adaptation}
Some studies on domain adaptation also focus on batch normalization. 
Li et al.~\cite{BN3} achieved a deep adaptation effect to the new domain by modulating the statistics from the source domain to the target domain in the batch normalization layers.
Chang et al.~\cite{BN4} train separate batch normalization layers for source and target domains. The batch normalization layers for the target domain are trained with pseudo-labels generated by existing methods~\cite{BN5,BN6}.
In the present work, we propose a technique different from that used in the earlier studies for the batch normalization layers.

\section{Problem Description}
\label{sec-setting}
In this section, we define one-shot DIL.
Let $\mathcal{X}(\subseteq \mathbb{R}^d)$ be an input space.
Let $\mathcal{Y}=\{1,\ldots,K\}$ be the set of classes, where $K\in\mathbb{N}$ is the number of classes.
Let $\mathcal{D}_{\mathrm{orig}}=\{(\bm{x}_n, y_n)\}^N_{n=1}\ (\bm{x}_n\in\mathcal{X}, y_n\in\mathcal{Y};\ n=1,\ldots,N)$ be a labeled dataset for classification,
where $\bm{x}_n\sim p_{\mathrm{orig}}(\bm{x}|y=k)(k=1,\ldots,K)$.
In this work, $f_{\bm{\theta}} : \mathcal{X}\to\mathcal{Y}$ parametrized by $\bm{\theta}\in\bm{\Theta}(\subseteq\mathbb{R}^p)$ denotes a neural network model.
We suppose that a neural network model $f_{\hat{\bm{\theta}}_{\mathrm{orig}}}(\hat{\bm{\theta}}_{\mathrm{orig}}\in\bm{\Theta})$ trained with~$\mathcal{D}_{\mathrm{orig}}$ is given.

Under this premise, we suppose that a new labeled dataset~$\mathcal{D}_{\mathrm{new}}$ is given.
We assume that $|\mathcal{D}_{\mathrm{new}}|=1$ and represent~$\mathcal{D}_{\mathrm{new}}=\{(\bm{x}_0, y_0)\}\ (\bm{x}_0\in\mathcal{X}, y_0\in\mathcal{Y})$.
The new sample $(\bm{x}_0, y_0)$ is based on the following assumptions:
\begin{itemize}
    \item The trained model $f_{\hat{\bm{\theta}}_{\mathrm{orig}}}$ misclassifies the new sample~$(\bm{x}_0, y_0)$.
    \item The sample $(\bm{x}_0, y_0)$ is not included in the original dataset~$\mathcal{D}_{\mathrm{orig}}$.
    \item $\bm{x}_0\sim p_{\mathrm{new}}(\bm{x}|y=y_0)$, where $p_{\mathrm{orig}}(\bm{x}|y=y_0)$ and $p_{\mathrm{new}}(\bm{x}|y=y_0)$ are not equivalent.
\end{itemize}

We define one-shot DIL as updating the trained model~$f_{\hat{\bm{\theta}}_{\mathrm{orig}}}$ with~$\mathcal{D}_{\mathrm{new}}$ to obtain a new model $f_{\hat{\bm{\theta}}_{\mathrm{new}}}(\hat{\bm{\theta}}_{\mathrm{new}}\in\bm{\Theta})$ that can correctly classify the inputs on the new domain $p_{\mathrm{new}}(\bm{x}|y=y_0)$. Note that the model is required to maintain the classification pattern on the original domain $p_{\mathrm{orig}}(\bm{x}|y=k)$ as much as possible. 
As in previous studies~\cite{GEM,DER}, we allow a subset~$\mathcal{M}$ of the examples in the original data $\mathcal{D}_{\mathrm{orig}}$ to be stored. We can use the subset~$\mathcal{M}$ in one-shot DIL to prevent the deterioration of the accuracy on the original domain.

For example, in Fig.~\ref{fig}, the original domain contains classes other than trucks. The new domain contains only trucks.
The given truck image corresponds to $\bm{x}_0$, and the label~$y_0$ represents the automobile class.
We expect that the new model~$f_{\hat{\bm{\theta}}_{\mathrm{new}}}$ will correctly classify the trucks as well as the other data.
Most previous studies about DIL assumed that the domains expand in all classes.\footnote{In one example of DIL experiments, MNIST~\cite{MNIST} is set as the original domain, and the permuted MNIST~\cite{pMNIST}, which has the same labels as MNIST, is set as the new domain.}
Unlike them, we assume that the domain increases in only one class because the new dataset contains only one sample.
\section{One-Shot DIL with Existing DIL Methods}
\subsection{Existing DIL Methods}
\label{secewcgem}
One-shot DIL is similar to general DIL, except that the number of training samples is limited in one-shot DIL.
Therefore, we applied the existing methods for general DIL to our one-shot DIL setting.
Some studies proposed DIL methods with mechanisms to maintain accuracy on the original domain even after training on the new domain;
some DIL methods require a replay buffer, while others do not.

GEM~\cite{GEM} is a well-known method that requires a replay buffer.
In GEM, the loss gradient of the new dataset is converted into a vector whose angle with the loss gradient of the original dataset is within 90°.
By updating the weights with this new vector, we can guarantee that the loss of the original dataset does not increase locally.

In contrast, EWC~\cite{EWC} is a method that does not require a replay buffer.
In EWC, a regularization term is added to the cross-entropy loss to reduce the distance between the updated weights and the weights obtained from the previous training.
This regularization term is intended to prevent the neural network models from forgetting the knowledge gained from past training.
\subsection{One-Shot DIL with EWC and GEM}
\label{secprob}
\begin{table}[t]
    \centering
        \caption{Test accuracy on the original and new domains when training 1000 new domain samples or one new domain sample.}
        \label{tab0}
        \begin{tabular}{ccccc}\toprule
            \multirow{2}{*}{model}                                         & \multicolumn{2}{c}{settings} & \multicolumn{2}{c}{test accuracy}                                           \\\cmidrule(lr){2-3}\cmidrule(lr){4-5}
                                                                           & methods                         & $|\mathcal{D}_{\mathrm{new}}|$     & $p_{\mathrm{new}}$ & $p_{\mathrm{orig}}$ \\\midrule
            \multirow{6}{*}{$f_{\hat{\bm{\theta}}_{\mathrm{new}}}$} & \multirow{2}{*}{CE}             & 1000                               & 0.9410             & 0.9420             \\
                                                                           &                                 & 1                                  & 0.5765             & 0.4936             \\\cmidrule{2-5}
                                                                           & \multirow{2}{*}{CE+EWC}         & 1000                               & 0.9390             & 0.9434             \\
                                                                           &                                 & 1                                  & 0.5955             & 0.4647             \\\cmidrule{2-5}
                                                                           & \multirow{2}{*}{CE+GEM}         & 1000                               & 0.9960             & 0.9306             \\
                                                                           &                                 & 1                                  &              0.7315      &  0.9071             \\\midrule
            $f_{\hat{\bm{\theta}}_{\mathrm{orig}}}$                  & \multicolumn{2}{c}{\raisebox{-0.5mm}{―}}        & \raisebox{-0.5mm}{0.6940}                             & \raisebox{-0.5mm}{0.9550}                                  \\\bottomrule
        \end{tabular}
    \vspace{-0.3cm}
\end{table}
EWC and GEM have mechanisms to improve accuracy on the new domain while maintaining accuracy on the original domain, as discussed in Section~\ref{secewcgem}.
Therefore, it is natural to expect that the same results can be achieved in one-shot DIL as well.
However, in experiments simulating one-shot DIL, we could not maintain the accuracy on the original domain even by using EWC and GEM.
Further, one-shot DIL was intended to improve the accuracy on the new domain, in one-shot DIL with EWC and GEM, the accuracy of the new domain was worsened.

The failure in one-shot DIL is illustrated in Table~\ref{tab0}, which summarizes the test accuracy\footnote{We explain the dataset to calculate the test accuracy in Section~\ref{secdataset}} on the new domain $p_{\mathrm{new}}(\bm{x}|y=y_0)$ and the original domain $p_{\mathrm{orig}}(\bm{x}|y=k)(k=1,\ldots,K)$. We simulated DIL for the example given in Fig.~\ref{fig}.
We used ResNet18~\cite{ResNet} as a model for classification.
In the experiments, we first trained the model by the training dataset of CIFAR10~\cite{CIFAR} without trucks as the original data $\mathcal{D}_{\mathrm{orig}}$.\footnote{The setup of the first training is explained in Section~\ref{secex}.}
We then updated this trained model with trucks as the new data $\mathcal{D}_{\mathrm{new}}$ in two settings, namely $|\mathcal{D}_{\mathrm{new}}|=1000$ and $|\mathcal{D}_{\mathrm{new}}|=1$.
The latter is the one-shot DIL setting and the former is an easier setting than the one-shot DIL regarding the larger number of new data. For the new data $\mathcal{D}_{\mathrm{new}}$, we randomly selected 1000 samples or one sample from truck images in CIFAR10.
As the subset~$\mathcal{M}$, we randomly selected 1000 samples from $\mathcal{D}_{\mathrm{orig}}$. We concatenated them with $\mathcal{D}_{\mathrm{new}}$ to update the model.
In the $|\mathcal{D}_{\mathrm{new}}|=1000$ setting, we updated the trained model by Adam~\cite{Adam} with the learning rate $10^{-5}$ and the number of iterations 100.
The training setup for the $|\mathcal{D}_{\mathrm{new}}|=1$ setting is explained in Section~\ref{secex}. In the $|\mathcal{D}_{\mathrm{new}}|=1$ setting, we resampled the new data with data augmentation~\cite{DataAug} at each step of the optimization so that each mini-batch contains the same number of samples from both the new domain and the original domain.\footnote{How to resample the new data shows at Line 4-9 in Alg.~\ref{alg1}. Batch normalization is performed as the normal method in the experiments described in Section~\ref{secprob}.}
CE denotes the case where we performed optimization with simple cross-entropy loss.
CE+EWC and CE+GEM are the same as CE except that EWC or GEM is applied.

When $|\mathcal{D}_{\mathrm{new}}|=1$, the accuracy on both domains decreases after training with $\mathcal{D}_{\mathrm{new}}$, in the case of CE and CE+EWC.
This deterioration on both domains is a problem specific to one-shot DIL as the deterioration on the new domain does not occur when $|\mathcal{D}_{\mathrm{new}}|=1000$.
In addition, existing DIL methods other than EWC and GEM are expected to show the same trend given the failure of CE.
We carefully investigated the cause of accuracy deterioration in one-shot DIL.
In the following section, we explain why this deterioration occurs and how to solve this problem.
\section{Method for One-Shot DIL}
\subsection{Why EWC and GEM Fail in One-Shot DIL}
\label{secshift}
\begin{figure}[t]
     \vspace{-0.5cm}
    \centerline{\includegraphics[height=4.9cm,width=7.3cm]{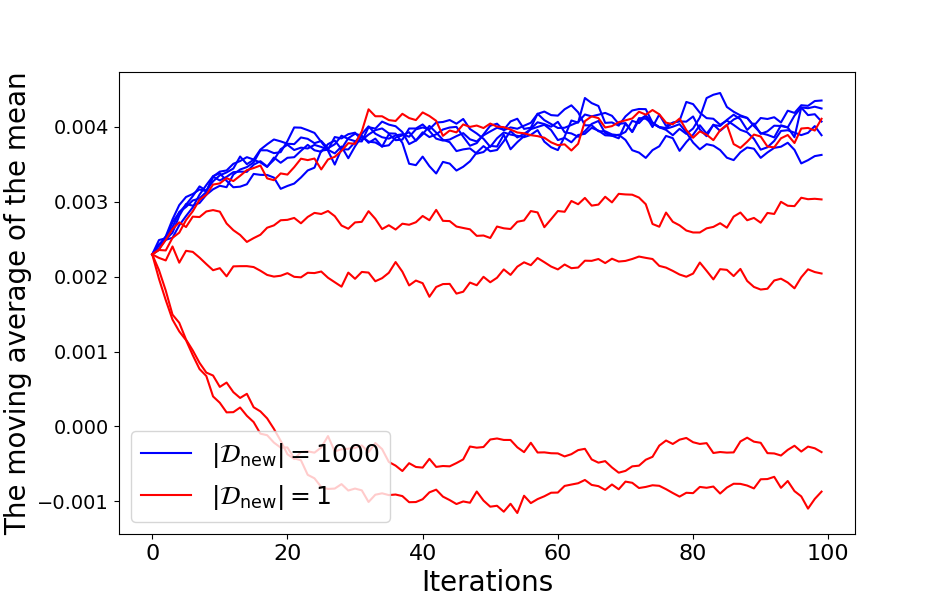}}
    \caption{Transition of the moving averages of the mean at the batch normalization layer closest to the input layer in ResNet18~\cite{ResNet}.
        In training, the moving averages of the statistics in the batch normalization layer are repeatedly updated at every forward propagation.
        Therefore, we accumulated the moving averages of the statistics at every forward propagation and plotted their sequences.
        Since the input is normalized in parallel for each channel in the batch normalization layer, we computed the average for the channels.}
    \label{figmean}
    \vspace{-0.5cm}
\end{figure}

\begin{figure}[t]
    \centerline{\includegraphics[height=4.9cm,width=7.3cm]{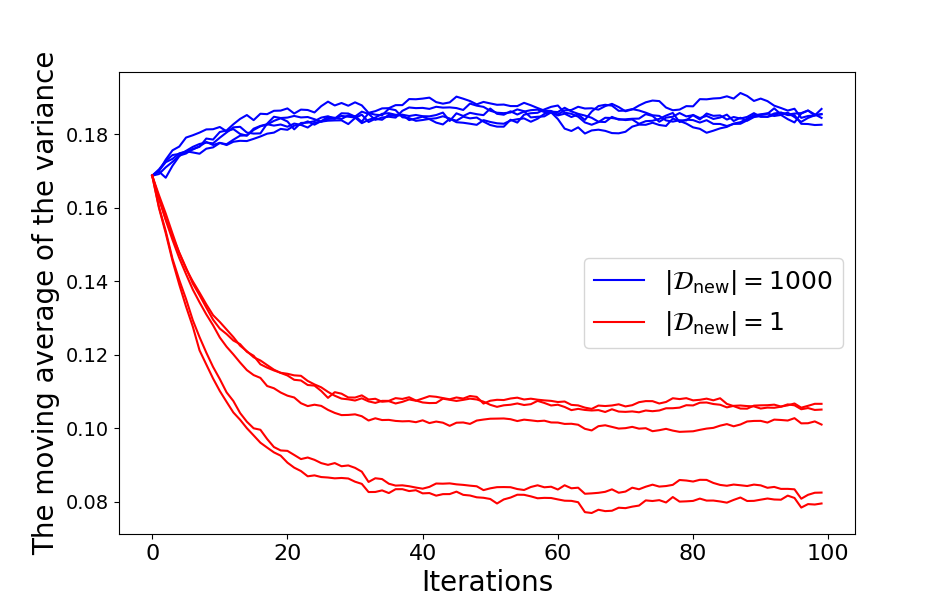}}
    \caption{Transition of the moving averages of the variance at the batch normalization layer closest to the input layer in ResNet18~\cite{ResNet}. The settings of the plot are the same as those shown in Fig.~\ref{figmean}.}
    \label{figvar}
    \vspace{-0.3cm}
\end{figure}

As discussed in Section~\ref{secprob}, we cannot achieve sufficient accuracy even when we apply EWC or GEM directly to one-shot DIL.
We carefully analyzed the cause of this problem and discovered that the cause lies in the statistics in the batch normalization layers.
Figures \ref{figmean} and \ref{figvar} show the transitions of the moving averages of the mean and variance of a batch normalization layer in ResNet18. We chose an example in CE.
We ran five trials with different $\mathcal{D}_{\mathrm{new}}$ for each of the $|\mathcal{D}_{\mathrm{new}}|=1000$ and $|\mathcal{D}_{\mathrm{new}}|=1$ settings.
We plotted all results of the five trials as the five lines in the figures.

In Fig.~\ref{figmean}, the moving average of the mean shifts in more widely different directions over the five trials in $|\mathcal{D}_{\mathrm{new}}|=1$ than in $|\mathcal{D}_{\mathrm{new}}|=1000$.
When $|\mathcal{D}_{\mathrm{new}}|=1$, for some trials, a positive moving average is obtained, while for others, a negative moving average is obtained.
This variation is caused by the fact that the mean value is calculated under a strong influence of the single input $\bm{x}_0$. 
The mean value heavily influenced by the single input $\bm{x}_0$ causes the trainable weights and biases in the batch normalization layers to be updated in unexpected directions during training.
Furthermore, in inference, the moving average that is heavily influenced by the single input $\bm{x}_0$ degrades the generalization ability.

In Fig.~\ref{figvar}, the moving average of the variance shifts in the opposite direction between $|\mathcal{D}_{\mathrm{new}}|=1000$ and $|\mathcal{D}_{\mathrm{new}}|=1$.
DIL is expected to increase the variance by adding the new domain.
As shown in Fig.~\ref{figvar}, the moving average of the variance shifts as expected when $|\mathcal{D}_{\mathrm{new}}|=1000$. 
However, when $|\mathcal{D}_{\mathrm{new}}|=1$, the moving average of the variance decreases.
This discrepancy is due to a lack of diversity in the new data in one-shot DIL.
This results in a lack of variance in the mini-batches, in turn causing the trainable weights and biases in the batch normalization layers to be updated in unexpected directions during training.
Furthermore, in inference, the much smaller moving average of variance causes misclassification because the outputs of the batch normalization layers become much larger than the inputs.
The results summarized in Table~\ref{tab0} reflect the above problems.
Therefore, we propose to modify the handling of the statistics in the batch normalization layers for one-shot DIL.
\subsection{Modifying Batch Normalization Statistics}
\label{secbatch-norm}
As explained in Section~\ref{secshift}, the shifted statistics in the batch normalization layers degrade the accuracy on the new and original domains under the one-shot DIL setting.
Therefore, we modify the statistics as follows:
\begin{itemize}
    \item We do not update the moving averages of the statistics and we fix them to the values they had before training the new domain.
    \item We use these constant moving averages for batch normalization when calculating forward and backward propagation during training.
    \item We also use the constant moving averages in inference.
\end{itemize}
Note that the above settings can only be used after training on the original data $\mathcal{D}_{\mathrm{orig}}$. We cannot apply them for non-continual one-shot learning.
The above improvement avoids the unexpected shift of the statistics shown in Section~\ref{secshift}. Therefore, we can prevent the mean heavily influenced by the single input $\bm{x}_0$ from degrading the generalization ability and eliminate much larger outputs of the batch normalization layers due to the lack of variance in the mini-batches. Our improvement can be applied to existing DIL methods and does not counteract the effectiveness of the existing methods.

\subsection{Algorithm}
\label{secalg}
\begin{figure}[t]
    \removelatexerror
    \begin{algorithm}[H]
        \caption{Our one-shot DIL algorithm with improved batch normalization.}
        \label{alg1}
        \KwInput{$\mathcal{M}$, $\mathcal{D}_{\mathrm{new}}=\{(\bm{x}_0, y_0)\}$, $f_{\hat{\bm{\theta}}_{\mathrm{orig}}}$,
        $\mathcal{L}$:~loss~function,~$\nu$:~learning rate, $\mathrm{transform}$:~the~random~operator~of~augmentation,
        $\mathrm{judge} : \bm{\Theta}\to\{\mathrm{True}, \mathrm{False}\}$:~the~judgment for termination of iteration}
        \KwOutput{$f_{\hat{\bm{\theta}}_{\mathrm{new}}}$}

        $\bm{\theta}_0\leftarrow \hat{\bm{\theta}}_{\mathrm{orig}}$

        $t \leftarrow 0$

        \While{$\mathrm{judge}(\bm{\theta}_t)=\mathrm{False}$}{
        Sample a mini-batch $\mathcal{B}$ from $\mathcal{M}$.

        \For{$m=1,2,\ldots,M$}{
            $\bm{z}_{m}\leftarrow  \mathrm{transform}(\bm{x}_0)$
        }

        $\mathcal{C}\leftarrow \{(\bm{z}_{1}, y_0), (\bm{z}_{2}, y_0), \ldots, (\bm{z}_{M}, y_0)\}$

        $\mathcal{B}'\leftarrow \mathcal{B}\cup\mathcal{C}$

        Compute the forward and backward propagation of the loss $\mathcal{L}(\bm{\theta}_t;\mathcal{B}')$ with the fixed statistics explained in Section~\ref{secbatch-norm}.

        $\bm{\theta}_{t+1}\leftarrow \bm{\theta}_t-\nu \nabla_{\bm{\theta}_t}\mathcal{L}(\bm{\theta}_t; \mathcal{B}')$

        $t \leftarrow t+1$

        }
        $\hat{\bm{\theta}}_{\mathrm{new}}\leftarrow \bm{\theta}_{t}$

        \Return $f_{\hat{\bm{\theta}}_{\mathrm{new}}}$
    \end{algorithm}
    \vspace{-0.3cm}
\end{figure}

Algorithm~\ref{alg1} shows the details of our algorithm for one-shot DIL.
We assume that the neural network model $f_{\hat{\bm{\theta}}_{\mathrm{orig}}}$ trained on the original data $\mathcal{D}_{\mathrm{orig}}$ is given, as in Section~\ref{sec-setting}.
In one-shot DIL, the new domain $p_{\mathrm{new}}(\bm{x}|y=y_0)$ becomes the training target in addition to the original domain $p_{\mathrm{orig}}(\bm{x}|y=k)(k=1,\ldots,K)$.
However, we have only one sample on the new domain, denoted as $(\bm{x}_0, y_0)$. This makes training on the new domain difficult.
Therefore, we apply data augmentation~\cite{DataAug} to $\bm{x}_0$, as described in Lines 5-8 in Alg.~\ref{alg1}.
We replicate the new data $\bm{x}_0$ into $M$ samples ($M\in\mathbb{N}$) and transform them randomly and differently in each iteration. The transformation methods are described in Section~\ref{secsetup}.
The vectors~$\bm{z}_1,\bm{z}_2,\ldots,\bm{z}_M\in\mathcal{X}$ denote the new samples obtained by transforming $\bm{x}_0$.
We assume that the transformed samples~$\bm{z}_1,\bm{z}_2,\ldots,\bm{z}_M$ belong to the same class $y_0$ as $\bm{x}_0$.
However, if all the samples in the mini-batches belong to the class $y_0$, updating $\bm{\theta}$ with these mini-batches may result in a model that classifies all the inputs into $y_0$.
To avoid this problem, we concatenate the mini-batch~$\mathcal{B}$, which is the mini-batch of $\mathcal{M}$, with $\mathcal{C}=\{(\bm{z}_m, y_0)\}^M_{m=1}$ (Line 9).

Then, in forward and backward propagation (Line 10), we perform batch normalization with the fixed statistics explained in Section~\ref{secbatch-norm}.
The moving averages of the statistics are not updated.
We continue the training iteration as long as a function ``$\mathrm{judge}$'' returns ``$\mathrm{False}$'' and stop it when ``$\mathrm{judge}$'' returns ``$\mathrm{True}$'' (Line 3).
We define ``$\mathrm{judge}$'' as the function that returns ``$\mathrm{True}$'' if the element corresponding to $y_0$ in the softmax output of $f_{\bm{\theta}}(\bm{x}_0)$ is greater than a threshold $\delta\in[0, 1]$.
``$\mathrm{judge}$'' returns “False” if it is less than $\delta$.

When using EWC in general DIL, the cross entropy is computed only with new data~\cite{EWC}.
In one-shot DIL, however, we compute the cross entropy with $\mathcal{B}'$ in Alg.~\ref{alg1} to prevent the trained model from classifying all the inputs into $y_0$. The Fisher information needed for EWC is computed with $\mathcal{M}$.
For GEM, we compute the loss for the new domain with $\mathcal{B}'$, similar to EWC. The loss for the original domain is computed with a mini-batch $\mathcal{B}$ randomly sampled from $\mathcal{M}$.
\section{Experiments}
\label{secex}
To investigate the effectiveness of modifying the batch normalization, we performed experiments on the image datasets MNIST~\cite{MNIST}, CIFAR10~\cite{CIFAR}, and RESISC45~\cite{RESISC}.
We trained models in one-shot DIL with the following two settings and calculated the test accuracy on both the new domain $p_{\mathrm{new}}(\bm{x}|y=y_0)$ and the original domain $p_{\mathrm{orig}}(\bm{x}|y=k)(k=1,\ldots,K)$.
\begin{description}
    \item[\textbf{updated-stats}: ]\hspace{1.2cm} In the batch normalization layers, the statistics are calculated from the inputs as usual. Simultaneously, the moving averages of the statistics are updated.
    \item[\textbf{fixed-stats}: ]\hspace{0.7cm} As explained in Section~\ref{secbatch-norm}, the batch normalization layers constantly use the moving averages obtained before training the new domain. These moving averages are not updated.
\end{description}
Note that there is no difference other than the batch normalization layers.
In both settings, we sampled the mini-batches per Lines 4-9 in Alg.~\ref{alg1}. We tried two cases where $|\mathcal{B}|=|\mathcal{C}|=32$ and $|\mathcal{B}|=63, |\mathcal{C}|=1$. For each setup, we tried CE, CE+EWC, and CE+GEM as in Section~\ref{secprob}.
\subsection{Datasets}
\label{secdataset}
MNIST~\cite{MNIST} and CIFAR10~\cite{CIFAR} classify images into 10 classes, while RESISC45~\cite{RESISC} classifies images into 45 classes. 
For each dataset, we selected two classes, denoted by~$C_1$ and~$C_2$, and merged the samples in the two classes. We assumed that the merged samples have the same label $y_0$. We also assumed that $\bm{x}\sim p_{\mathrm{new}}(\bm{x}|y=y_0)$ if $\bm{x}$ belongs to $C_1$ and $\bm{x}\sim p_{\mathrm{orig}}(\bm{x}|y=y_0)$ if $\bm{x}$ belongs to $C_2$.
After this merger, we had only one less classes than the original MNIST, CIFAR10, and RESISC45.
Table~\ref{tab1} lists the settings of~$C_1$ and~$C_2$ in our experiments with MNIST, CIFAR10, and RESISC45. We adopted two settings per dataset, which are named Set 1 and Set 2. For example, Set 1 of CIFAR10 is shown in Fig.~\ref{fig}.

For each setting in Table~\ref{tab1}, we randomly divided the dataset assigned as the original domain into a training set, a validation set, and a test set. We regarded the training set as the original data $\mathcal{D}_{\mathrm{orig}}$. We trained a model $f_{\bm{\theta}}$ with $\mathcal{D}_{\mathrm{orig}}$ and got $f_{\hat{\bm{\theta}}_{\mathrm{orig}}}$. Moreover, we also randomly divided the dataset assigned as the new domain into a training set, a validation set, and a test set. We randomly selected 10 samples misclassified by~$f_{\hat{\bm{\theta}}_{\mathrm{orig}}}$ from the training set and performed one-shot DIL assuming that each sample was the new data $\mathcal{D}_{\mathrm{new}}=\{(\bm{x}_0, y_0)\}$. Note that one-shot DIL was run 10 times according to the number of samples selected. For the subset $\mathcal{M}$, we randomly selected 1000 samples from $\mathcal{D}_{\mathrm{orig}}$. As explained in Section~\ref{sec-setting}, we updated $f_{\hat{\bm{\theta}}_{\mathrm{orig}}}$ with $\mathcal{D}_{\mathrm{new}}$ and~$\mathcal{M}$ through Alg.~\ref{alg1} to obtain $f_{\hat{\bm{\theta}}_{\mathrm{new}}}$. Over the 10 trials, we calculated the median and standard deviation of the accuracy. The accuracy was calculated using the test sets on the original domain and the new domain.
\begin{table}[t]
    \centering
        \caption{Settings of the original and new domains in $y_0$.}
        \label{tab1}
        \vspace{-0.3cm}
        \begin{tabular}{ccccc}\toprule
                  &\multicolumn{2}{c}{Set 1} & \multicolumn{2}{c}{Set 2}   \\\cmidrule(lr){2-3}\cmidrule(lr){4-5}
            &  $C_1$ & $C_2$ & $C_1$ & $C_2$\\ \midrule
            MNIST & ``9" & ``8" & ``1'' & ``0'' \\
            CIFAR10 & ``truck'' & ``automobile" & ``dog" & ``cat"       \\
            RESISC45 & ``airplane'' & ``airport'' & ``overpass'' & ``intersection'' \\\bottomrule
        \end{tabular}
    \vspace{-0.3cm}
\end{table}
\begin{table*}
\centering
            \caption{Test accuracy on the new and original domains in the updated-stats or fixed-stats (MNIST)}
            \label{tab1-}
            \begin{tabular}{cccccccc}\toprule
                \multirow{2}{*}{model}                                         & \multicolumn{3}{c}{settings} & \multicolumn{2}{c}{test accuracy (Set 1)} & \multicolumn{2}{c}{test accuracy (Set 2)}                                          \\\cmidrule(lr){2-4}\cmidrule(lr){5-6}\cmidrule(lr){7-8}
                                                                               & \multicolumn{1}{c}{methods} &        batch size                &  batch norm                    & $p_{\mathrm{new}}$ & $p_{\mathrm{orig}}$ & $p_{\mathrm{new}}$ & $p_{\mathrm{orig}}$ \\\midrule
                \multirow{12}{*}{$f_{\hat{\bm{\theta}}_{\mathrm{new}}}$} & \multirow{4}{*}{CE} & \multirow{2}{*}{$|\mathcal{B}|=|\mathcal{C}|=32$}             & updated-stats                           & 0.7671$\pm$0.1880             & 0.9940$\pm$0.0203 &  0.8639$\pm$0.0763 & 0.9933$\pm$0.0049           \\
                                                                               &               &                  & fixed-stats                            & 0.8826$\pm$0.0715             & 0.9904$\pm$0.0046 & 0.9313$\pm$0.1016 & 0.9931$\pm$0.0056            \\\cmidrule{3-8}
                                                                               &              & \multirow{2}{*}{$|\mathcal{B}|=63, |\mathcal{C}|=1$} & updated-stats & 0.7522$\pm$0.2107  & 0.9959$\pm$0.2806 & 0.9110$\pm$0.0692 & 0.9946$\pm$0.0005 \\
                                                                               &              & & fixed-stats & 0.8251$\pm$0.0832 & 0.9921$\pm$0.0026 & 0.9194$\pm$0.1089 & 0.9936$\pm$0.0010 \\\cmidrule{2-8}
                                                                               & \multirow{4}{*}{CE+EWC} & \multirow{2}{*}{$|\mathcal{B}|=|\mathcal{C}|=32$}        & updated-stats                           & 0.8414$\pm$0.1454            & 0.9953$\pm$0.0068 & 0.8656$\pm$0.1085 & 0.9931$\pm$0.0377          \\
                                                                               &                    &             & fixed-stats                           & 0.8930$\pm$0.0611            & 0.9903$\pm$0.0050 & 0.9256$\pm$0.0972 & 0.9932$\pm$0.0055            \\\cmidrule{3-8}
                                                                               &               & \multirow{2}{*}{$|\mathcal{B}|=63, |\mathcal{C}|=1$} & updated-stats & 0.7190$\pm$0.1416 & 0.9959$\pm$0.0013 & 0.8863$\pm$0.0847 & 0.9941$\pm$0.0004 \\
                                                                               &              & & fixed-stats & 0.8746$\pm$0.0778 & 0.9917$\pm$0.0022 & 0.9167$\pm$0.0783 & 0.9937$\pm$0.0026 \\\cmidrule{2-8}
                                                                               & \multirow{4}{*}{CE+GEM}   & \multirow{2}{*}{$|\mathcal{B}|=|\mathcal{C}|=32$}      & updated-stats        &      0.7800$\pm$0.1108       &    0.9958$\pm$0.0043  & 0.8819$\pm$0.1008  &     0.9936$\pm$0.0015       \\
                                                                               &                       &          & fixed-stats         &        0.8677$\pm$0.0688     &   0.9909$\pm$0.0035 & 0.9198$\pm$0.0968   &    0.9937$\pm$0.0026        \\\cmidrule{3-8}
                                                                               &               & \multirow{2}{*}{$|\mathcal{B}|=63, |\mathcal{C}|=1$} & updated-stats &  0.7507$\pm$0.1017  & 0.9957$\pm$0.0019  &     0.8903$\pm$0.0751  &   0.9941$\pm$0.0011    \\
                                                                                &              & & fixed-stats &   0.8593$\pm$0.0863 & 0.9928$\pm$0.0047 &  0.9022$\pm$0.0676  & 0.9941$\pm$0.0008   \\\midrule
                $f_{\hat{\bm{\theta}}_{\mathrm{orig}}}$                  & \multicolumn{3}{c}{\raisebox{-0.5mm}{―}}        & \raisebox{-0.5mm}{0.1021}                             & \raisebox{-0.5mm}{0.9962} & \raisebox{-0.5mm}{0.0555} & \raisebox{-0.5mm}{0.9951}                                  \\\bottomrule
            \end{tabular}
\end{table*}
\begin{table*}
\centering
            \caption{Test accuracy on the new and original domains in the updated-stats or fixed-stats (CIFAR10)}
            \label{tab2-}
            \begin{tabular}{cccccccc}\toprule
                \multirow{2}{*}{model}                                         & \multicolumn{3}{c}{settings} & \multicolumn{2}{c}{test accuracy (Set 1)} &      \multicolumn{2}{c}{test accuracy (Set 2)}                                      \\\cmidrule(lr){2-4}\cmidrule(lr){5-6}\cmidrule(lr){7-8}
                                                                               & methods            &   batch size          & batch norm                          & $p_{\mathrm{new}}$ & $p_{\mathrm{orig}}$ & $p_{\mathrm{new}}$ & $p_{\mathrm{orig}}$ \\\midrule
                \multirow{12}{*}{$f_{\hat{\bm{\theta}}_{\mathrm{new}}}$} & \multirow{4}{*}{CE}     & \multirow{2}{*}{$|\mathcal{B}|=|\mathcal{C}|=32$}          & updated-stats                           & 0.5765$\pm$0.2297            & 0.4936$\pm$0.1847 & 0.7595$\pm$0.3077 & 0.9363$\pm$0.2308             \\
                                                                               &               &                  & fixed-stats                           & 0.9595$\pm$0.0430             & 0.9335$\pm$0.0440 & 0.9335$\pm$0.0257 & 0.9528$\pm$0.0267            \\\cmidrule{3-8}
                                                                                &              & \multirow{2}{*}{$|\mathcal{B}|=63, |\mathcal{C}|=1$} & updated-stats & 0.6895$\pm$0.1466 & 0.9413$\pm$0.0054 & 0.7740$\pm$0.0771 & 0.9544$\pm$0.0051 \\
                                                                                &              & & fixed-stats & 0.9260$\pm$0.0425 & 0.9424$\pm$0.0091 & 0.9050$\pm$0.0110 & 0.9580$\pm$0.0023 \\\cmidrule{2-8}
                                                                               & \multirow{4}{*}{CE+EWC}    & \multirow{2}{*}{$|\mathcal{B}|=|\mathcal{C}|=32$}       & updated-stats                           & 0.5955$\pm$0.2614             & 0.4647$\pm$0.2202 & 0.7665$\pm$0.2721 & 0.9360$\pm$0.2210             \\
                                                                               &                   &              & fixed-stats                           & 0.9420$\pm$0.0586             & 0.9332$\pm$0.0192 & 0.9330$\pm$0.0241 & 0.9522$\pm$0.0259            \\\cmidrule{3-8}
                                                                               &              & \multirow{2}{*}{$|\mathcal{B}|=63, |\mathcal{C}|=1$} & updated-stats & 0.7815$\pm$0.1008 & 0.9394$\pm$0.0070 & 0.7305$\pm$0.0526 & 0.9580$\pm$0.0057 \\
                                                                               &              & & fixed-stats & 0.9375$\pm$0.0456 & 0.9414$\pm$0.0164 & 0.9035$\pm$0.0260 & 0.9587$\pm$0.0204 \\\cmidrule{2-8}
                                                                               & \multirow{4}{*}{CE+GEM}  & \multirow{2}{*}{$|\mathcal{B}|=|\mathcal{C}|=32$}        & updated-stats                &         0.7315$\pm$0.2287     &  0.9071$\pm$0.2512   &  0.8081$\pm$0.1010  &     0.9556$\pm$0.0313          \\
                                                                               &                  &               & fixed-stats                           &  0.9485$\pm$0.0417          &  0.9404$\pm$0.0250  & 0.9240$\pm$0.0255   &   0.9556$\pm$0.0201         \\\cmidrule{3-8}
                                                                               &              & \multirow{2}{*}{$|\mathcal{B}|=63, |\mathcal{C}|=1$} & updated-stats &  0.8015$\pm$0.0901    & 0.9399$\pm$0.0062  &  0.8075$\pm$0.0663   &  0.9532$\pm$0.0051  \\
                                                                               &              &                                                   & fixed-stats & 0.9220$\pm$0.0452   &  0.9431$\pm$0.0137 &   0.8985$\pm$0.0223 & 0.9584$\pm$0.0041   \\\midrule
                $f_{\hat{\bm{\theta}}_{\mathrm{orig}}}$                  & \multicolumn{3}{c}{\raisebox{-0.5mm}{―}}        & \raisebox{-0.5mm}{0.6940}                             & \raisebox{-0.5mm}{0.9550} & \raisebox{-0.5mm}{0.7800} & \raisebox{-0.5mm}{0.9672}                                 \\\bottomrule
            \end{tabular}
\end{table*}
\begin{table*}
\centering
            \caption{Test accuracy on the new and original domains in the updated-stats or fixed-stats (RESISC45)}
            \label{tab3-}
            \begin{tabular}{cccccccc}\toprule
                \multirow{2}{*}{model}                                         & \multicolumn{3}{c}{settings} & \multicolumn{2}{c}{test accuracy (Set 1)} &      \multicolumn{2}{c}{test accuracy (Set 2)}                                      \\\cmidrule(lr){2-4}\cmidrule(lr){5-6}\cmidrule(lr){7-8}
                                                                               & methods            &   batch size          & batch norm                          & $p_{\mathrm{new}}$ & $p_{\mathrm{orig}}$ & $p_{\mathrm{new}}$ & $p_{\mathrm{orig}}$ \\\midrule
                \multirow{12}{*}{$f_{\hat{\bm{\theta}}_{\mathrm{new}}}$} & \multirow{4}{*}{CE}     & \multirow{2}{*}{$|\mathcal{B}|=|\mathcal{C}|=32$}          & updated-stats       & 0.4250$\pm$0.1217         & 0.9482$\pm$0.0096  & 0.1500$\pm$0.0584  & 0.9505$\pm$0.0105         \\
                                                                               &               &                  & fixed-stats               &   0.8250$\pm$0.0862       & 0.8912$\pm$0.0712 & 0.7179$\pm$0.1591  &     0.9265$\pm$0.0455      \\\cmidrule{3-8}
                                                                                &              & \multirow{2}{*}{$|\mathcal{B}|=63, |\mathcal{C}|=1$} & updated-stats & 0.2750$\pm$0.1067 & 0.9515$\pm$0.0117 & 0.0429$\pm$0.0450 & 0.9535$\pm$0.0025 \\
                                                                                &              & & fixed-stats & 0.7179$\pm$0.1433 & 0.9403$\pm$0.0117 & 0.5250$\pm$0.1475 & 0.9388$\pm$0.0172 \\\cmidrule{2-8}
                                                                               & \multirow{4}{*}{CE+EWC}    & \multirow{2}{*}{$|\mathcal{B}|=|\mathcal{C}|=32$}       & updated-stats        &       0.3893$\pm$0.1416       &  0.9502$\pm$0.0166 &  0.1393$\pm$0.0776 & 0.9481$\pm$0.0118             \\
                                                                               &                   &              & fixed-stats                           &       0.8214$\pm$0.0868     &  0.8879$\pm$0.0728  &  0.7179$\pm$0.1591  & 0.9265$\pm$0.0456           \\\cmidrule{3-8}
                                                                               &              & \multirow{2}{*}{$|\mathcal{B}|=63, |\mathcal{C}|=1$} & updated-stats & 0.2750$\pm$0.0637 & 0.9539$\pm$0.0023 & 0.0500$\pm$0.0421 & 0.9543$\pm$0.0022 \\
                                                                               &              & & fixed-stats & 0.7179$\pm$0.1433 & 0.9402$\pm$0.0117 & 0.5250$\pm$0.1475 & 0.9389$\pm$0.0171 \\\cmidrule{2-8}
                                                                               & \multirow{4}{*}{CE+GEM}  & \multirow{2}{*}{$|\mathcal{B}|=|\mathcal{C}|=32$}        & updated-stats     &   0.3643$\pm$0.1081       &     0.9478$\pm$0.0083     &   0.1429$\pm$0.0616      &    0.9499$\pm$0.0108     \\
                                                                               &                  &               & fixed-stats                           &    0.7786$\pm$0.1020       &   0.9294$\pm$0.0549   &    0.5786$\pm$0.1512  &     0.9356$\pm$0.0217          \\\cmidrule{3-8}
                                                                               &              & \multirow{2}{*}{$|\mathcal{B}|=63, |\mathcal{C}|=1$} & updated-stats &     0.3321$\pm$0.1187  &     0.9546$\pm$0.0231    &     0.0929$\pm$0.0463   &    0.9563$\pm$0.0006   \\
                                                                               &              & & fixed-stats &  0.7393$\pm$0.1559   &   0.9410$\pm$0.0124  &  0.4714$\pm$0.1187   &   0.9425$\pm$0.0113  \\\midrule
                $f_{\hat{\bm{\theta}}_{\mathrm{orig}}}$                  & \multicolumn{3}{c}{\raisebox{-0.5mm}{―}}        & \raisebox{-0.5mm}{0.1071}                             & \raisebox{-0.5mm}{0.9580} & \raisebox{-0.5mm}{0.0429} & \raisebox{-0.5mm}{0.9584}                                 \\\bottomrule
            \end{tabular}
\end{table*}
\subsection{Setup for Model Training}
\label{secsetup}
For the transformations from $\bm{x}_0$ to $\bm{z}_1,\ldots,\bm{z}_M$ in~Alg.~\ref{alg1}, we used RandomRotation, RandomResizedCrop, RandomAffine, and RandomPerspective, which were implemented in Torchvision~\cite{Torchvision}.
For ``$\mathrm{judge}$'' in Alg.~\ref{alg1}, we set $\delta=0.99$.
The neural network architecture used in the experiments was ResNet18~\cite{ResNet} for all datasets.
In the first training with only the original data $\mathcal{D}_{\mathrm{orig}}$, we used SGD~\cite{SGD} as the optimizer. The learning rate was scheduled by cosine annealing~\cite{CosineAnnealing} from 0.1 to 0.0, and there were 200 epochs.
In one-shot DIL, we used Adam~\cite{Adam} as the optimizer. We calculated the accuracy of the validation set of the original domain\footnote{We did not use the validation set of the new domain because we assume that the number of samples on the new domain is only one in one-shot DIL.} for the fixed learning rate settings of $10^{-8}, 10^{-7}, 10^{-6}, 10^{-5}, 10^{-4}, 10^{-3}, 10^{-2}$, and $10^{-1}$.
Then, we selected the settings that satisfied the constraint that the training iterations were terminated (``$\mathrm{judge}$'' returns ``$\mathrm{True}$'') within 100 iterations.
We adopted the learning rate with the highest accuracy among the selected settings.
\subsection{Results}
Table~\ref{tab1-},~\ref{tab2-}, and~\ref{tab3-} list the Set 1 and 2 results of MNIST, CIFAR10, and RESISC45, respectively.
All tables list the test accuracy on the new domain $p_{\mathrm{new}}(\bm{x}|y=y_0)$ and original domain $p_{\mathrm{orig}}(\bm{x}|y=k)$.
As explained in Section~\ref{secdataset}, we calculated the median and standard deviation of the accuracies over 10 trials, with 10 different samples. We show them in the tables (median$\pm$standard deviation).

In the case of MNIST, the updated-stats improved the accuracy on the new domain through one-shot DIL; however, compared to the updated-stats, the fixed-stats achieved higher accuracy on the new domain.  
From Table~\ref{tab1-}, a comparison of the fixed-stats with the updated-stats shows that the accuracy on the new domain increased by 0.8\%-15\% while maintaining the reduction in accuracy on the original domain below 0.6\%.
Although there is a trade-off between the accuracy on the new domain and that on the original domain, the decrease in the accuracy on the original domain is much smaller than the increase in the accuracy on the new domain.

In the case of CIFAR10, some settings of the updated-stats degraded the accuracy on both the new and original domains through one-shot DIL.
Thus, we can say that CIFAR10 is a more challenging dataset for one-shot DIL than MNIST.
However, the fixed-stats avoided the deterioration of accuracy and improved the accuracy on the new domain.
From Table~\ref{tab2-}, a comparison of the fixed-stats with the updated-stats shows that the accuracy on the new domain increased by 9\%-38\%, and the accuracy on the original domain increased by 0\%-46\%.
The fixed-stats improved the accuracy not only on the new domain but also on the original domain in the experiments with CIFAR10. There is no example where the fixed-stats is worse than the updated-stats in Table~\ref{tab2-}.

RESISC45 results show similar trends to MNIST.
From Table~\ref{tab3-}, a comparison of the fixed-stats with the updated-stats shows that the accuracy on the new domain increased by 37\%-57\% while maintaining the reduction in accuracy on the original domain below 7\%. Similar to MNIST, the decrease in the accuracy on the original domain is much smaller than the increase in the accuracy on the new domain.

Note that the difference between the updated-stats and the fixed-stats lies only in the statistics of the batch normalization layers. These results show that good accuracy can be achieved just by modifying the batch normalization statistics. Moreover, from the results of CE, we can predict that other DIL methods with cross-entropy loss show the same trend. Therefore, we set these results as a baseline for one-shot DIL.
\section{Conclusion}
In this study, we examined one-shot DIL and improved the statistics in the batch normalization layers for one-shot DIL.
The usual batch normalization in one-shot DIL leads to a variation in the moving average of the mean and a reduction in the moving average of the variance, which negatively affects accuracy.
Therefore, we proposed to normalize the inputs with the moving averages of the statistics obtained from training the original domain and not to update these moving averages when training the new domain.
By the proposed method, one-shot DIL caused an increase in the accuracy on the new domain and prevented severe degradation of the accuracy on the original domain.
Our method can be applied to existing DIL methods. Therefore, we hope that the combinations of our method and the existing DIL methods are used in many tasks requiring one-shot DIL.
This study contributes to many cases where samples from new domains are rarely observed over time, but the model must adapt to the new domains, such as self-driving and anomaly detection in a production line.
\input{main.bbl}

\bibliographystyle{IEEEtran}
\end{document}

%% file: main.bbl
% Generated by IEEEtran.bst, version: 1.12 (2007/01/11)